%% file: main.tex
\documentclass{article}
\usepackage{graphicx}
\usepackage{blindtext} 
\usepackage{subfig}
\usepackage{microtype}
\usepackage{natbib, url}
\usepackage{amsmath}
\usepackage{float}
\usepackage{authblk}  

\input{header}  
\hypersetup{pdfborder = {0 0 0.5 [3 3]}, colorlinks = true, linkcolor = BrightRed, citecolor = SkyBlue}

\usepackage[margin=1in]{geometry}
\usepackage{tikz}
\usepackage{setspace}
\usepackage[font=small,labelfont=bf]{caption}

\title{Probit Monotone BART}
\author[1]{Jared D. Fisher}
\affil[1]{Department of Statistics, Brigham Young University}

\date{August 29, 2025}

\begin{document}
\maketitle

\begin{abstract}
    Bayesian Additive Regression Trees (BART) of \cite{chipman2010bart} has proven to be a powerful tool for nonparametric modeling and prediction. Monotone BART \citep{chipman2022mbart} is a recent development that allows BART to be more precise in estimating monotonic functions. We further these developments by proposing probit monotone BART, which allows the monotone BART framework to estimate conditional mean functions when the outcome variable is binary. 
\end{abstract}

\section{Introduction}

Since it's advent in \cite{chipman2010bart}, Bayesian Additive Regression Trees, or BART, has become a major player in the worlds of Bayesian inference, machine learning, and causal inference \citep{hill2020bayesian}. 
BART's success is due in part to the regularization of tree depth, with the modeling belief that trees should be smaller/simpler than not. Using this piece of structure, when it's a good assumption, allows for powerful modeling. 

Likewise, if we have good reason to believe that relationships between quantitative covariates and the outcome are monotonic, our models would benefit from implementing that. \cite{chipman2022mbart} introduce BART with monotonic constraints, allowing monotone BART to estimate monotonic functions better than standard BART. However, their implementation assumes the outcome of interest is continuous and that the additive errors are normally distributed. 

In this paper, we introduce an implementation of monotone BART for binary outcomes. This is done with a probit link, implemented using the ideas of normal latent variables/data augmentation from \cite{albert1993bayesian}, akin to how \cite{chipman2010bart} created the probit-variant of original BART.

The paper proceeds as follows. We first review the original, probit, and monotone BART models from the literature. In Section \ref{sec:pmbart} we propose probit monotone BART, including the model set up and some details on the code implementation. In Section \ref{sec:sim} we demonstrate the proposed method with a simulation study. Section \ref{sec:conclusion} concludes.

\section{Existing BART Methods}

Before we proposal our model in Section \ref{sec:pmbart}, we first define original BART as well as the two variants of BART that our model proposal uses as cornerstones.

\subsection{Original BART}
\label{sec:bart}

Here, we summarize the construction of the  original BART model of \cite{chipman2010bart}. BART models a continuous outcome $Y$ as sum-of-trees model $f$ plus Gaussian errors $\epsilon$.  The conditional expectation of $Y$ given covariates $\bm{x}$ is modeled with the sum of $m$ regreesion trees $g$, where both the tree structure and leaf node values are regularized by Bayesian priors, as follows.  
\begin{align}
    Y_i &= f(x_i) + \epsilon_i, \quad 
    \epsilon_i \sim N(0,\sigma^2) \\
    f(x) &= \sum_{j=1}^m g(x; T_j, M_j)
\end{align}
Each regression tree $g$ follows the tree definition of \cite{chipman1998bayesiancart}, defined by $T_j$ and $M_j$. The tree structure $T_j$ partitions the domain  of $\bm{x}$ while $M_j$ is a vector of values $\{\mu_1,\mu_2,...,\mu_{b_j}\}$ that the function takes on within each cell of the partition, or equivalently, the value of each of the $b_j$ leaves/terminal nodes of tree $j$. The prior on the error variance and each of the priors for the regression trees are assumed independent, and the priors for each terminal node value $\mu_{\ell j}$ are independent conditional on the tree structure $T_j$:
\begin{align}
    p((T_1,M_1),...,(T_m,M_m),\sigma) &= \prod_j \left[p(T_j) \prod_\ell^{b_j} p(\mu_{\ell j}|T_j) \right]p(\sigma)
\end{align}

The prior for each tree structure/partition $T_j$ follows the guidance of \cite{chipman1998bayesiancart}, where there is uniform probability across which covariates are chosen to cut the partition/split the tree and uniform probability across cut points for the chosen covariate. The unique aspect though is the prior probability for tree depth $d$: $\alpha(1+d)^\beta$. The authors recommend $\alpha = 0.95$ and $\beta = 2$ as hyperparameter values that behave reasonably across many different scenarios. 

The original implementation shifted and rescaled $Y$ so that the training set ranged from $-0.5$ to 0.5, and given the tree structure, the leaf node values have normal priors that place probability over all of the space of observed outcomes. 
\begin{align}
\mu_{\ell j}|T_j  &\sim N(0,\sigma^2_\mu) \label{eq:pbart_prior_mu} \\
\sigma_\mu &= 0.5/(k\sqrt{m}) 
\end{align}
where the variance depends on the number of trees $m$ and the shrinkage parameter $k$.   $k = 2$ is the recommended default and typically they suggest $m=50$ or $m=200$ trees.

Lastly, the error standard deviation is given a conjugate inverse chi-square prior
\begin{align}
    \sigma^2 \sim \frac{\nu \lambda}{\chi^2_\nu}
\end{align}
    where the hyperparameters are chosen based on a ``rough data-based overestimate $\hat{\sigma}$ of $\sigma$''. We defer further details to the original paper.

\subsection{Probit BART}\label{sec:pbart}

Here we briefly discuss how the probit BART model differs from standard BART, both of which are from \cite{chipman2010bart}. Let $Y_i$ now be a binary outcome of interest. For observations $i=1,...,n$, let
        \begin{align}
            P[Y_i = 1| \bm{x}_i] &= \Phi[G(\bm{x}_i)+ c] \label{eq:probitlik} \\
            G(\bm{x}) &= \sum_{j=1}^m g(\bm{x}; T_j, M_j) 
        \end{align}
    where $c=\Phi^{-1}(\bar{y})$ is an offset (akin to centering the data by subtracting off the sample mean) and $G$ is a sum of $m$ regression trees $g$. However as $Y$ is now binary/Bernoulli instead of normal/Gaussian, there is no $\sigma$. 
    
    As with standard BART, each regression tree $g$ follows the tree definition of \cite{chipman1998bayesiancart}, defined by $T_j$ and $M_j$, and the priors for the regression trees are assumed independent, and the priors for each terminal node value $\mu_{\ell j}$ are independent conditional on the tree structure $T_j$.  
        \begin{align}
    P((T_1,M_1),...,(T_m,M_m)) &= \prod_{j=1}^m \left[p(T_j)p(M_j|T_j) \right]\\
    p(M_j|T_j) &= \prod_\ell^{b_j} p(\mu_{\ell j}|T_j)  
        \end{align}
    where $\ell$ indexes the $b_j$ different terminal nodes of tree $j$. 
    
    Given the tree structure, the leaf node values again have normal priors
    \begin{align}
    \mu_{\ell j}|T_j &\sim N(0,\sigma^2_\mu)   \\
    \sigma_\mu &= 3/(k\sqrt{m}) 
    \end{align}
    but now the prior standard deviation $\sigma_\mu$ is chosen with $3$ in the numerator instead of 0.5 ``such that G(x) will with high probability be in the interval (-3.0,3.0).'' This variance depends on the number of trees $m$ and the shrinkage parameter $k$. $k = 2$ is the recommended default and typically $m=50$ or $m=200$. 

    Computationally, the posterior sampling of this probit regression is done with the data augmentation with normal latent variables as per \cite{albert1993bayesian}. We replace Equation \ref{eq:probitlik} by introducing latent variables $Z_i$ where
    \begin{align}
        Y_i &= \begin{cases} 
        1, Z_i > 0\\
        0, Z_i \le 0
        \end{cases} 
        \label{z_intro}
        \\
        Z_i &\sim N(G(x_i),1) \label{z_dist}
    \end{align}
    such that posterior sampling of the trees proceeds much the same as before, but the likelihood used by BART is based on $Z$ instead of $Y$, and there are new posterior draws of $Z_i$ from 
    \begin{align}
        Z_i|(Y_i = 1) &\sim max\{N(G(x),1),0\}\\
        Z_i|(Y_i = 0) &\sim min\{N(G(x),1),0\}. \label{z0}
    \end{align}

\subsection{Monotone BART}

Montonic BART though differs from the original BART with its monotonic-constrained prior \citep[Section~3]{chipman2022mbart}.
Let 
\begin{equation}    
C=\left\{ (T,M) : g(x;T,M) \text{ is monotone for each }x_i \in S\right\}
\end{equation}
where $S$ is a subset of the coordinates of $x\in R^n$. Monotonicity is incorporated by constraining the prior of $P(M_j|T_j) = \prod_i p(\mu_{ij}|T_j)$ to have support only over $C$: 
\begin{align}
    P(M_j|T_j) \propto \left[\prod_i^{b_j} p(\mu{ij}|T_j)\right] \chi_C(T_j,Mj)
\end{align}
where $b_j$ is the number of bottom/terminal nodes of $T_j$ and $\chi_C(.) = 1$ on $C$ and 0 otherwise.

The prior on tree shape/size, $p(T_j)$, largely follows original BART. The variables for trees to split on, and the values they split at, are chosen with uniform probability.  While  $\alpha = .95$ and $\beta = 2$ are typically suggested, they suggest $\alpha = .25$ and $\beta = .8$ as these hyperparameter values yield posterior trees that are comparable in size to trees from standard BART \citep[Section~4.3]{chipman2022mbart}.
 
The prior on leaf node values, $p(\mu_{ij}|T_j)$, is calibrated to have larger variance than the standard BART prior  \citep[Section~3.3]{chipman2022mbart}. They let
\begin{align}
(\mu_{ij}|T_j) \sim N(\mu_\mu, c^2\sigma^2_\mu)
\end{align}
where $c^2 = \frac{\pi}{\pi-1} \approx 1.4669$. Akin to the choices of $\alpha$ and $\beta$, this increase in variance is to allow the marginal distributions of the leaf nodes of short trees to have variance close to those from original BART. 
$\mu_\mu$ and $\sigma_\mu$ are chosen so that the prior on $E(Y|x)$ has substantial probability on the interval ($y_{min},y_{max}$). For 
\begin{align}
    m \mu_\mu - k\sqrt{m}\sigma_\mu = y_{min}\\
    m \mu_\mu + k\sqrt{m}\sigma_\mu = y_{max}
\end{align}
they again recommend centering and scaling the $Y_i$ to have a minimum of -0.5 and maximum 0.5 like before, such that $\mu_\mu = 0$ and $\sigma_\mu = 0.5/(k\sqrt{m})$. 
$k=2$ is suggested so that $E(Y|x)$  a 95\% prior probability on the observed range of $Y$. Lastly, they suggest choosing $m$, the number of trees, in advance as a tuning parameter, with $m\ge 50$ for prediction \citep[Section~3.4]{chipman2022mbart} . 

MCMC sampling of the constrained posterior of monotone BART is fairly involved, and we defer further details to their paper.

\section{Probit Monotone BART}
\label{sec:pmbart}

\subsection{The Model}
We propose incorporating the normal latent random variables from \cite{albert1993bayesian}, as used by \cite{chipman2010bart}, into the monotone BART framework of \cite{chipman2022mbart}. 
We thus make two main changes to probit BART: constrain/truncate the priors on the leaf nodes to yield monotonic functions of $\bm{x}$ and increase the variance of the prior by $c^2$. 
Likewise, this implies two main changes to monotone BART: introduce the normal latent variables and change the likelihood of the outcome from normal to Bernoulli, thus removing $\sigma$ from the model. Here we enumerate all parts of the proposed model, reiterating some of the Equations from the previous section. 

For $Y_i \sim Bernoulli(p_i)$, following probit BART, let
 \begin{align}
            p_i &= P[Y_i = 1| \bm{x}_i] = \Phi[G(\bm{x}_i)+ c] \\
            G(\bm{x}) &= \sum_{j=1}^m g(\bm{x}; T_j, M_j) 
        \end{align}
 where $c=\Phi^{-1}(\bar{y})$ is an offset and $G$ is a sum of $m$ regression trees $g$.  The priors for the regression tree parameters are assumed independent, and the priors for each terminal node value $\mu_{\ell j}$ are independent conditional on the tree structure $T_j$.  
        \begin{align}
    P((T_1,M_1),...,(T_m,M_m)) &= \prod_{j=1}^m \left[p(T_j)p(M_j|T_j) \right].
        \end{align}
        The prior on the tree structures $p(T_j)$ follows the standard BART protocol, with uniform probability on the splitting variables, uniform probability on the splitting locations, and regularizes the depth/size of the tree by setting the prior probably of tree depth $d$ to be $\alpha(1+d)^\beta$, for $\alpha \in (0,1), \beta \in [0,\infty)$.

    Like in monotone BART, our monotonic constraints are imposed through the prior on each tree's vector of leaf node values, 
    \begin{align}
            p(M_j|T_j) &\propto \left[\prod_\ell^{b_j} p(\mu_{\ell j}|T_j)\right] \chi_C(T_j,M_j)
    \end{align}
where $\chi$ is an indicator function equal to 1 when $(T_j,M_j) \in C$, and 0 otherwise, and  $\ell$ indexes the $b_j$ different terminal nodes of tree $j$.  $C$ is the set of all parameters that create a monotonic tree, that is, borrowing Equation 3.2 from \cite{chipman2022mbart}, 
    \begin{equation}
        C = \{(T,M): g(\bm{x};T,M) \text{ is monotone for each } x_{p} \in S\}
    \end{equation}
    where $S$ is a subset of $\{1,...,P\}$, i.e. the coordinates of $\bm{x} \in R^P$.  In short, the terminal node values $\mu_{\ell j}$ are sampled from distributions truncated by their neighboring nodes such that $g(\bm{x};T,M)$ is nondecreasing in the desired covariates. 
    The priors on the individual leaf node values given the tree structure, $p(\mu_{\ell j}|T_j)$, are normal priors with inflated variance 
    \begin{align}
            (\mu_{\ell j}|T_j) &\sim N(0, c^2\sigma^2_\mu)
    \end{align}
    where $c^2 = \frac{\pi}{\pi - 1} \approx 1.467$. The purpose of scaling the variance by $c^2$ is to allow the marginal variance of each $\mu_{\ell j}$ to be $\sigma^2_\mu$ (in the case of small trees), as the monotonicity constraints impose prior dependence between the $\mu_{\ell j}$, $\ell \in \{1,...,b_j\}$. Lastly, we follow probit BART in setting
    \begin{align}
    \sigma_\mu &= 3/(k\sqrt{m}).
    \end{align}

   For hyperparameters, we choose the standard number of trees $m=200$, $k=2$ as suggested in both papers, and $\alpha = .25$ and $\beta = .8$   following \cite{chipman2022mbart}.

\subsection{Computation}

 Following  the general pattern of the probit BART components from the \func{gbart} function in the \pkg{BART} R package \citep{sparapani2021nonparametric}, we add normal latent variables to the main function from monotone BART package \citep{chipman2022mbart}, following the ideas of \cite{albert1993bayesian} and \cite{chipman2010bart}.  These additions follow Equations \ref{z_intro} to \ref{z0}, reproduced here: 
    \begin{align*}
        Y_i &= \begin{cases} 
        1, Z_i > 0\\
        0, Z_i \le 0
        \end{cases} 
         \\
        Z_i &\sim N(G(x_i),1)  \\
        Z_i|(Y_i = 1) &\sim max\{N(G(x),1),0\}\\
        Z_i|(Y_i = 0) &\sim min\{N(G(x),1),0\}. 
    \end{align*}

 Our proposed probit monotone BART function, \func{probit\_monbart}, is available as part of our modified version of the \citep{chipman2022mbart}'s R package, \pkg{mBARTprobit}, available on Github \citep{mBARTprobit}.



\section{Simulation}
\label{sec:sim}

Here we briefly present a simulation study to demonstrate the benefit of monotonic constraints in nonparametric probit regression. For $i=1,...,500,$ let $Y_i \sim Bernoulli(p_i)$ with $p_i = \Phi[f(x_i)]$, representing a probit regression on $x$. Let $x_i\sim U(-3,3)$ and 
$$
f(x) = 
\begin{cases}
0.2x, & \text{if } x < 0 \\
x, & \text{if } x \geq 0
\end{cases}
$$
such that the binary outcomes are generated from a Bernoulli distribution whose probabilities follow a piecewise, but monotonic, probit regression curve. 

\begin{figure}[htpb]
    \centering
    \includegraphics[width=0.8\linewidth]{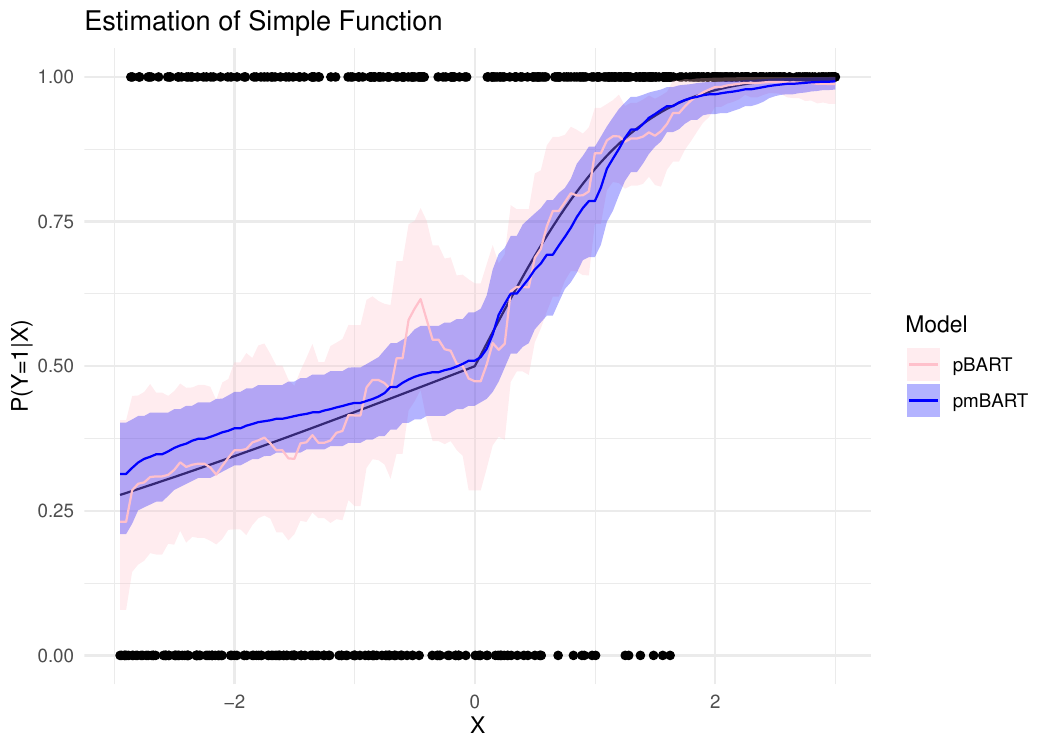}
    \caption{Comparison of probit BART to probit monotone BART for fitting a monotone probit regression. Observed data points and the true mean function are in black. Probit BART, or pBART, is shown in pink. Probit monotone BART, or pmBART, is shown in blue. The solid lines are the posterior mean curves and the transparent region shows the 90\% pointwise posterior credible region, for each model.}
    \label{fig:simple}
\end{figure}

We fit both probit BART and probit monotone BART with default settings. The posterior mean curve and 90\% pointwise credible bands for each model are shown in Figure \ref{fig:simple}. We see that when the mean function is truly monotonic, imposing monotonic constraints improves both the estimation and its precise, as shown by how close the blue curve (probit monotone BART's posterior mean) is to the true curve in black, as well as how narrow probit monotone BART's credible region is compared to that of probit BART. 

 \section{Conclusion}
 \label{sec:conclusion}
We present probit monotone BART, an important addition to the suite of BART tools that now allows monotone BART to estimate conditional expectations for binary outcome variables. An early version of the posterior sampler is available on Github \citep{mBARTprobit}. 

\section*{Acknowledgments}
We thank Rodney Sparapani for his helpful suggestion.

 \newpage
\bibliographystyle{apalike}
\bibliography{references}

\end{document}

%% file: header.tex
\usepackage{amsmath, amssymb, amsthm}
\usepackage{algorithm}
\usepackage{algpseudocode}
\usepackage{amsmath, amssymb, amsthm}
\usepackage{float, graphicx, multirow,subcaption, setspace}
\usepackage{comment}
\usepackage{url}
\usepackage{enumitem}
\usepackage{tikz}
\usepackage{bm, bbm}
\usetikzlibrary{shapes.geometric, positioning}
\usetikzlibrary{quotes, angles}
\usepackage{rotating}
\usepackage{hyperref}
\usepackage{natbib}
\usepackage[normalem]{ulem}

\definecolor{SkyBlue}{RGB}{14, 118, 188}
\definecolor{BrightRed}{RGB}{223, 82, 78}
\definecolor{Green638}{RGB}{165,255,118} 
\definecolor{FoamGreen}{RGB}{25, 200, 150}

\definecolor{RevisedRed}{RGB}{238, 51, 119}

\newlist{assumenum}{enumerate}{1}
\setlist[assumenum]{label=\textbf{A\theassumenumi}, ref=A\theassumenumi}

\newlist{assumenumx}{enumerate}{1}
\setlist[assumenumx]{label=\textbf{Ax\theassumenumxi}, ref=Ax\theassumenumxi}

\theoremstyle{plain}

\theoremstyle{definition}

\newtheorem{ex}{Example}

\theoremstyle{plain}

\newcommand{\pkg}[1]{\textit{#1}}
\newcommand{\func}[1]{\texttt{#1}}